\documentclass[lettersize,journal]{IEEEtran}

\usepackage{amsmath,amsfonts}
\usepackage{algorithmic}
\usepackage{algorithm}
\usepackage{array}
\usepackage[caption=false,font=normalsize,labelfont=sf,textfont=sf, position=top]{subfig}
\usepackage{textcomp}
\usepackage{stfloats}
\usepackage{url}
\usepackage{verbatim}
\usepackage{graphicx, wrapfig}
\usepackage{cite}
\usepackage[dvipsnames]{xcolor}
\usepackage{multirow}
\usepackage{hyperref}
\usepackage{todonotes}
\usepackage[english]{babel}
\usepackage{tikz}
\usepackage{amssymb}
\usetikzlibrary{shapes, arrows}
\addto\extrasenglish{%
	% ...

	% ...
}

\usepackage{lineno}
\begin{document}

\title{Iterative Cluster Harvesting for Wafer Map Defect Patterns}
%

\author{Alina~Pleli*, Simon~Baeuerle*, Michel~Janus, Jonas~Barth, Ralf~Mikut, Hendrik~P.~A.~Lensch
	\thanks{A.~Pleli and Hendrik~P.~A.~Lensch are with the department of Computer Graphics, University of Tuebingen, D-72076 Tuebingen, Germany. (e-mail: alina.pleli@de.bosch.com) (*Alina~Pleli and Simon~Baeuerle are co-first authors.) (corresponding author: S.~Baeuerle.)}
	\thanks{S.~Baeuerle and R.~Mikut are with the Institute for Automation and Applied Informatics (IAI), Karlsruhe	Institute of Technology (KIT), D-76344 Eggenstein-Leopoldshafen, Germany.}
	\thanks{A.~Pleli, S.~Baeuerle, M.~Janus and J.~Barth are with the Robert Bosch GmbH, D-72762 Reutlingen, Germany.}
}

\markboth{}
{Pleli, Baeuerle \MakeLowercase{\textit{et al.}}: Iterative cluster harvesting for wafer map defect patterns}


\maketitle

\begin{abstract}
Unsupervised clustering of wafer map defect patterns is challenging because the appearance of certain defect patterns varies significantly. This includes changing shape, location, density, and rotation of the defect area on the wafer. We present a harvesting approach, which can cluster even challenging defect patterns of wafer maps well. Our approach makes use of a well-known, three-step procedure: feature extraction, dimension reduction, and clustering. The novelty in our approach lies in repeating dimensionality reduction and clustering iteratively while filtering out one cluster per iteration according to its silhouette score. This method leads to an improvement of clustering performance in general and is especially useful for difficult defect patterns. The low computational effort allows for a quick assessment of large datasets and can be used to support manual labeling efforts. We benchmark against related approaches from the literature and show improved results on a real-world industrial dataset.
\end{abstract}

\section{Introduction}
\label{sec:introduction}
\IEEEPARstart{S}{emiconductor} manufacturing is a fast-growing branch of industry, which is necessary to further advance developments, e.g. in the
field of automotive electrification and many other applications. Worldwide, the demand for powerful and complex semiconductor chips is increasing. Individual processing steps are precisely controlled
and errors should be detected as quickly and reliably as possible. Because of the large number of chips produced in a manufacturing facility, an automated error recognition is needed and can be supported by machine learning. Positions and types of defects found at different production steps of
individual chips on a wafer can be displayed as an image, which is called wafer map. Clustering these wafer map images based on the type(s) of observed defect patterns should help to identify relevant defect patterns on the wafer maps.
These patterns offer insights that can be used for root-cause analysis of defects. For example, if there is a line at certain points on the wafer, a device may have scratched the wafer during a production step.  \\
The semiconductor manufacturing process is highly complex and variable. New defects can constantly occur where labeled data is not available and even for known defect patterns, labeling is a time-intensive task requiring expert knowledge.
Many other approaches have been published to classify defect patterns. This includes classical supervised approaches~\cite{Nakazawa_similarity_search_hamming}, transfer learning approaches using neural networks with similarity search~\cite{CNN_transfer_wafermaps, similarity_wm_resnet_knn} and novel approaches using autoencoders~\cite{cluster_wm_varae_santos, VarDeepClustering_wafermaps}. These usually require extensive and well-tuned training of the network architecture. In addition, for supervised classification training the set of error patterns must be specified beforehand and it relies on precise labeling.

We present an unsupervised machine-learning approach that does not require expert knowledge, is flexible for new, unknown defect patterns while at the same time simplifying the training setup. Our approach relies on iteratively removing clusters from 
the dataset based on the silhouette score~\cite{Silhouette_paper} and redefining the feature space to maximize the separation between wafer-defect patterns in each iteration.
This approach enables fast clustering on high-dimensional image data by feature extraction and dimensionality reduction. 
Our approach further alleviates the information loss inherent in these procedures by iteratively removing clusters and redefining the feature space. Additionally, the method filters out some defect patterns from the dataset, which are not noticeable in other clustering approaches, unless you apply methods like noise clustering.

We investigate the performance of our method using a real-world industrial wafer map dataset with expert labels for well-known
recurring defect patterns and compare it against state-of-the-art methods. On this dataset we show that more homogeneous
clusters are formed in comparison to a non-iterative procedure. 

In \autoref{sec:relatedwork} related approaches are presented to frame the proposed method in the area of conceptually and computationally similar methods.
The methodology of our proposed approach is described in detail in
\autoref{sec:methodology}, and the method setup and data sets used are introduced in
\autoref{sec:experimentalsetup}. With this, the results are presented and discussed in
\autoref{sec:results}-\autoref{sec:conclusion}.
\section{Related Work}
\label{sec:relatedwork}
Gu\^{e}rin et al.~\cite{guerin_cnn_2018} analyzed a two-step approach to cluster images.
They used a convolutional neural network (CNN) as feature extractor followed by a clustering algorithm and analyzed different combinations of both.
They found that a combination of an Xception \cite{Xception_original_paper} feature extractor works well with agglomerative clustering (AC) \cite{Jain1988AlgorithmsFC, Rajalingam2011HierarchicalCA}.
Napoletano et al.~\cite{napoletano_anomaly_2018} investigated a three-step method with the application to images of nanofibrous materials. They use a ResNet-18 CNN as feature extractor together with a principal component analysis (PCA) and a k-means clustering. %

A variety of methods has been explored to perform clustering iteratively. An iterative Gaussian mixture model (GMM) clustering approach for scene images has been introduced by Doan et al.~\cite{iterative_clustering_gmmboost}, where the weights of the original dataset are iteratively computed. In this work, real scenes were evaluated along semantic axes (e.g., degree of naturalness of a scene, degree of verticality, and degree of openness). 
Compared to Doan et al. the feature extraction in our method is interchangeable and modular. The extractor should not be specified on the dataset (generic) and feature extraction should be done separately from the dimensional reduction. 

Shorewala~\cite{iterative_clustering_kmeans} extend the k-means algorithm. They drop samples with a distance larger than two standard deviations to the cluster mean from their assigned cluster and collect those as outliers within a separate group. Ma~\cite{iterative_clustering_dbscan} proposed an extension of the Density-Based Spatial Clustering of Applications with Noise (DBSCAN) clustering method with regard to clusters of different sample densities. Lin et al.~\cite{iterative_clustering_GIC} developed a general iterative clustering approach via improving the proximity matrix and clusters of a base random forest.
These iterative clustering methods~\cite{iterative_clustering_kmeans, iterative_clustering_dbscan, iterative_clustering_GIC} were however not yet applied to image datasets.

Intarapanich et al.~\cite{intarapanich_iterative_2009} introduce an algorithm to cluster subpopulations in genotypic data.
They iteratively apply a combination of PCA and fuzzy c-means clustering~\cite{bezdek_pattern_1981} to the dataset.
With each iteration the dataset is split into two subsets until a stopping criterion is reached.
They argue, that this procedure can resolve subpopulations better, especially if the samples of some subpopulations share a common region in PCA space while other subpopulations are located further away.
This approach has been enhanced multiple times~\cite{limpiti_iterative_2011, limpiti_study_2011, limpiti_injclust:_2014, c._amornbunchornvej_improved_2012, chaichoompu_methodology_2017}.
Those modifications include e.g. a modified stopping criterion~\cite{limpiti_iterative_2011, limpiti_injclust:_2014} or the assignment of individuals to an outlier group if they form a cluster with very few members~\cite{chaichoompu_methodology_2017}.
These iterative approaches are all studied genotypic data.
Odong et al.~\cite{odong_improving_2013} use a PCA in combination with a clustering algorithm on genotypic data without iterations and report good results as compared to directly performing clustering on raw data.

The silhouette score~\cite{Silhouette_paper} computes the similarity of a point to its own cluster compared to other clusters and is mainly used to determine the number of clusters and generally to compare different cluster results~\cite{silhouette_score_cluster_quality, silhouette_score_cluster_evaluation}. Nidheesh et al. \cite{silhouette_score_hierarchical_clustering} utilized the silhouette score to cluster genomic data for cancer subtype discovery in a hierarchical bottom-up procedure. To the best of our knowledge, the silhouette score has not yet been used to automate iterative clustering.

Clustering on wafer maps has been heavily studied in recent years. Thus, different methods have been investigated for clustering wafer maps. Besides classical or variational autoencoders (AEs)~\cite{cluster_wm_ae_knn, VarDeepClustering_wafermaps, cluster_wm_varae_santos}, or Dirichlet process mixture model (DPMM)~\cite{cluster_wm_ae_dpmm}, classical clustering methods such as DBSCAN~\cite{clustering_wm_dbscan} are also explored.
Additionally, hierarchical clustering methods (HCM) have been explored for clustering wafer maps. HCM start with each data point as its own cluster and iteratively try to combine the two most similar clusters into one. Zhang et al.~\cite{Zhang2013AutomaticCO} detected abnormal patterns via extracting of a small number of features by using sparse regression. These features are used for a complete-link hierarchical clustering.
Lee et al.~\cite{Similiarity_Clustering_of_wafermaps} propose a similarity ranking method based on the Dirichlet process Gaussian mixture model (DPGMM) with HCM-based on symmetric Kullback-Leibler divergence (SKLD). The similarity ranking is conducted via the Jensen-Shannon divergence.
Wang et al.~\cite{Wang_HCM_FCM} combine fuzzy c-means (FCM) with hierarchical linkage, Santos et al.~\cite{Santos_average_linkage} used average linkage and Tulala et al.~\cite{tulala_unsupervised_2018} the Ward's minimum variance criterion. 
For similarity search the Hamming distance was used~\cite{Nakazawa_similarity_search_hamming, Yu_similarity_search_hamming}, Yu et al. ranked top-N wafer map defect patterns via Euclidean distance~\cite{Yu_similarity_search_large_datasets} and Hsu et al. employed weighted and modified Hausdorff distance~\cite{Similarity_matching_Hausdorff_distance}.
Similar to Gu\^{e}rin et al., Wang et al. use a pre-trained CNN to extract features from image data, which are then clustered by k-means~\cite{similarity_wm_resnet_knn}.
Some other approaches with transfer learning are proposed~\cite{transfer_wm_classification_bhatnagar, transfer_wm_densenet}.

Unsupervised methods are helpful to retrieve similar defect patterns from large unlabeled datasets. However they are still outperformed in terms of accuracy by supervised methods, because an unsupervised method is not optimized on this metric in contrast to a supervised method. We introduce an iterative approach, which improves clustering performance as compared to its non-iterative counterpart. 
\tikzstyle{processtiny} = [rectangle, minimum width=0.5cm, minimum height=0.5cm, text centered,  text width=2cm, draw=black, fill=yellow!30]
\tikzstyle{processtiny2} = [rectangle, minimum width=0.5cm, minimum height=0.5cm, text centered,  text width=3cm, draw=black, fill=green!30]
\tikzstyle{processtiny3} = [rectangle, minimum width=0.5cm, minimum height=0.5cm, text centered,  text width=2cm, draw=black, fill=orange!30]
\tikzstyle{processtiny4} = [rectangle, minimum width=0.5cm, minimum height=0.5cm, text centered,  text width=2.5cm, draw=black, fill=orange!30]
\tikzstyle{process} = [rectangle, minimum width=0.5cm, minimum height=0.5cm, text centered,  text width=2.5cm, draw=black, fill=yellow!30]
\tikzstyle{decision} = [rectangle, minimum width=0.5cm, minimum height=0.5cm, text centered, draw=black, fill=yellow!30]
\tikzstyle{io} = [rectangle, minimum width=0.5cm, minimum height=0.5cm, text centered, draw=black, fill=yellow!30]
\tikzstyle{startstop} = [rectangle, minimum width=0.5cm, minimum height=0.5cm,text centered,text width=2.5cm, draw=black, fill=gray!30]
\tikzstyle{startstop2} = [rectangle, minimum width=0.5cm, minimum height=0.5cm, text centered,  text width=4cm, draw=black, fill=yellow!30]
\tikzstyle{arrow} = [thick,->,>=stealth]

\section{Methodology}
\label{sec:methodology}
\label{sec:proposed_method}
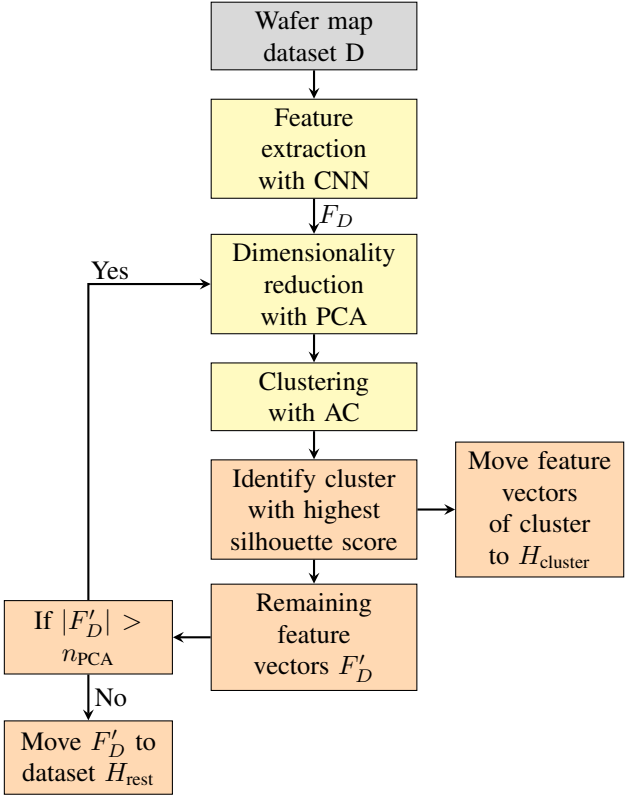
\begin{figure}
	\centering
	\begin{tikzpicture}
	
	\node (start) [startstop, yshift=-1cm] {Wafer map dataset D};
	\node (Feature) [process, below of=start,yshift=-.5cm] {Feature extraction with CNN};
	\draw [arrow] (start) -- (Feature);
	
	\node (dimred) [process, below of=Feature, yshift=-.8cm] {Dimensionality reduction with PCA};
	\draw [arrow] (Feature) -- node[yshift=0.0cm, xshift=+0.3cm] {$F_D$} (dimred);
	
	\node (clustering) [process, below of=dimred, yshift=-.5cm] {Clustering with AC};
	\draw [arrow] (dimred) -- (clustering);
	
	
		
	\node (harvest) [processtiny4, below of=clustering,yshift=-0.5cm, xshift=0cm ] {
		Identify cluster with highest silhouette score};
	\draw [arrow] (clustering) -- (harvest);
	
	\node (takebest) [processtiny3, right of=harvest,xshift=2.0cm ] {Move feature vectors of cluster to $H_{\text{cluster}}$};
	\draw [arrow] (harvest) -- (takebest);
	
	
	\node (remain) [processtiny4, below of=harvest,yshift=-0.7cm ] {Remaining feature vectors $F_D'$};
	\draw [arrow] (harvest) -- (remain);
	
	\node (while) [processtiny3, left of=remain,yshift=-0cm,xshift=-2.cm ] {If $|F_D'|>n_\text{PCA}$};
	\draw [arrow] (remain) -- (while);
	\draw [arrow] (while) |-  node[anchor=west,yshift=0.18cm, xshift=-0.1cm] {Yes} (dimred);
	
	\node (end) [processtiny3, below of=while,yshift=-0.6cm, xshift=0cm ] {Move  $F_D'$ to dataset $H_{\text{rest}}$};
	\draw [arrow] (while) --  node[yshift=0cm, xshift=0.3cm] {No} (end);

	
	

	%
	
	
	\end{tikzpicture}
	\caption{Iterative cluster harvesting scheme: CNN feature vectors $F_D$ are extracted for the dataset $D$ of wafer maps, and are processed with a dimensional 
		reduction (PCA in our case) and clustered (using AC in our case). Per iterative run, the cluster with the highest value for the silhouette 
		score is removed from $F_D$. Therefore in the subsequent iteration, the variances in the remaining dataset will lead to changes in the definitions 
		of the PCA components. This allows our method to further separate the remaining dataset and find more homogeneous clusters over successive iterations.}
		
	\label{fig:methodschema}
\end{figure}

We propose an iterative approach based on a three-step clustering process consisting of feature extraction and variance-based dimensionality reduction 
before clustering. By iteratively removing (or "harvesting") wafer maps already partitioned into good clusters and repeatedly reinitializing the 
variance-based dimensionality reduction on the shrinking dataset, the expectation is that variances in the feature space, that were not captured during the one-time application of the three-step procedure (one-time clustering or OTC), can be separated in successive iterations. We name this method \textbf{Iterative Cluster Harvesting (ICH)}.\\ 
In the following we use a pre-trained CNN for feature extraction, principal component analysis (PCA) as a well-known dimensionality reduction, which ranks features based 
on their variances in the given dataset, and Agglomerative Clustering (AC) to define the clusters in each iteration. However, we will show that the benefits of 
our iterative approach remain, even if one of the component methods in the three-step process is swapped out based on preference or expectation of better 
one-time performance for a given dataset.

In the following, we describe the procedure of our harvesting method step-by-step, analogous to the presented process flow shown in \autoref{fig:methodschema}. 

\subsection{Three-step clustering}
The core and start of the process flow is the aforementioned three-step clustering method. In the following, we describe each step separately, as it is also used in 
one-time clustering, before describing the ICH procedure and the changes in performance resulting from it. 

\subsubsection{Feature extraction}
The first step of the process flow is feature extraction with a CNN for the wafer map dataset $D$. For unsupervised clustering of images, the use of features from a pre-trained CNN is useful as this provides an abstract representation of images~\cite{Deep_learning_Overview_Schmidhuber}. %
In this paper, the choice of CNN architectures for feature extraction is not analyzed and interested readers are referred to~\cite{clustering_feature_extractor_analyse, guerin_cnn_2018}, where extensive experiments were conducted to identify the optimal feature extractor for different datasets. Thus, visually recognizable components of a wafer map are detected and can now be represented with a feature vector $f_i \in F_D$.
\subsubsection{Dimensionality reduction}
After the feature vectors are computed, the number of dimensions is still large enough to slow down and prohibit clustering from finding semantically relevant features.  We reduce this high number of dimensions further by performing a principal component analysis (PCA)~\cite{PCA_springer_book} to identify the features responsible for 
the largest variances in the dataset. It retains important and removes irrelevant dimensions. In this way, computational costs can be reduced significantly. For this 
method there is one freely selectable parameter: the number of principal components ($n_\text{PCA}$). The choice of this parameter can be dataset-dependent and 
problem-specific. \\
The dimensional reduction method itself is only one component within the clustering method of the proposed ICH approach in this paper and is not the focus of this study. For a mathematical description see~\cite{PCA_springer_book, Statistical_learning_Hastie}. We use the implementation of PCA from the Python package scikit-learn~\cite{scikit_learn}.
\subsubsection{Clustering}
Clustering of the resulting PCA feature space is conducted as a next step with the agglomerative clustering (AC) method. The algorithm is a hierarchical clustering algorithm in which each data point is considered as one cluster~\cite{Agglomerative}. From bottom to the top the clusters are agglomerated with Euclidean distance and the Ward's linkage. These are the default parameters of the scikit-learn~\cite{scikit_learn} implementation of the AC method. As a hyperparameter the number of expected clusters has to be determined. 
For a detailed description and comparison of the different methods see~\cite{Sasirekha2013AgglomerativeHC}. \\
A detailed investigation and optimization of the hyperparameters of the individual building blocks of the three-step clustering approach is not the 
objective of this work. Here, we also use the implementation of scikit-learn~\cite{scikit_learn}.
\subsection{Harvesting clusters with silhouette score}\label{sec:methodology_ich}
As shown in \autoref{fig:methodschema}, after the three steps of feature extraction, dimension reduction and clustering, the clusters are now sorted 
to determine the most suitable cluster to be harvested in the iteration. 
For this purpose, the silhouette score is calculated for each data point~\cite{Silhouette_paper}: 
\begin{align}
s_{i,j} = \frac{b-a}{\text{max}(a,b)}
\label{Formula: Silhouette}
\end{align}
with the mean distance $a$ between a data point $j$ and all other points in the cluster $c_i$ and the mean distance $b$ between a sample $j$ and all other data 
points in the next nearest cluster to $c_i$. In this way the silhouette score $s_{i,j}$ indicates how close one data point 
is to the other data points in its cluster $c_i$ in the PCA feature space and simultaneously calculates the distance to the next nearest cluster to it.\\
For each cluster $c_i$, we calculate the silhouette score s$_i$ as the mean of all $s_{i,j}$ for the included data points $j$. Thus, with this 
formula in \autoref{Formula: Silhouette}, dense clusters which are far away from other clusters are scored highly. Possible values for the silhouette 
score $s_{i}$ range from $-1$ to $1$, with an optimum value of $1$.\\
After calculating $s_{i}$ for all clusters we identify cluster $c_{s_{max}}$ with the maximum silhouette score value 
\begin{align}
s_{max} = \max_{i}~s_{i}.
\label{Formula: SilhouetteMax}
\end{align}
This cluster has the highest separability compared to the rest of the dataset and at the same time a high similarity of wafer maps within the cluster.\\
Consequently, the feature vectors $F_D \in c_{s_{max}}$ are moved to a dataset $H_{\text{cluster}}$.
Thus, one cluster is harvested to dataset $H_{\text{cluster}}$ per iteration based on the maximum of the silhouette scores in the specific iteration, 
and only the remaining CNN feature vectors $F_D'$ are used for the next iteration.\\
The sub-process of PCA and AC is then repeated, always harvesting one cluster in each iteration. As already mentioned, the PCA and the subsequent AC 
can be dominated at the start by the variance in the full data caused by individual samples and clusters whose feature vectors are very different from 
the rest of the dataset. Therefore the key aspect of our method is to successively remove data points from the dataset $D$ and repeat PCA and AC on the new dataset to successively focus on finer variances in the dataset. \\
This process is repeated until there are not enough wafer maps left in the dataset 
to sensibly perform a PCA. In our experiments we found if $\lvert F'_D\rvert > n_\text{PCA}$ to be a good rule-of-thumb for a stopping criterion. \\ 
Note that, when the stopping criterion is fulfilled, some feature vectors $F_D'$, which were never included in any dense and highly separated cluster, will 
remain unclustered. This dataset of feature vectors we denote as $H_{\text{rest}}$.

\subsection{Summary and extensions of method}
To summarize we restate the main features of our harvesting method: after the one-time feature extraction by a CNN, the dimension reduction and clustering 
is performed iteratively, harvesting per iteration the cluster with the highest $s_{i}$ in the PCA space during this iteration, thereby reducing the variance 
of the remaining dataset. 
Thus, at the end of iterative process, we have sorted most of the data points from the original dataset D into clusters in the dataset $H_{\text{cluster}}$, 
while separating out a dataset $H_{\text{rest}}$ of intricate wafer maps.

\subsubsection{Refining $H_{\text{cluster}}$ based on size}

During our studies we found it useful to further refine $H_{\text{cluster}}$. For our use case it is relevant that clusters are large enough. 
For example, if there is a cluster that contains less than 5 wafer maps, the significance of this defect pattern for root-cause analysis can be questioned.
So the clusters in $H_{\text{cluster}}$ are filtered by their size to a dataset $H_{\text{small}}$ and only large enough clusters are kept in $H_{\text{cluster}}$. The number of required 
wafer maps $n_{min}$ in a cluster can be adjusted depending on the underlying use case and the user's own definition of what is the minimum size of a cluster.
Such a filtering based on cluster size has been used in a similar fashion on genotypic data by Chaichoompu et al.~\cite{chaichoompu_methodology_2017}.

\subsubsection{Full assignment of wafer maps to $H_{\text{cluster}}$}
Both due to the existence of $H_{\text{rest}}$ and the extension that creates $H_{\text{small}}$, not all wafer maps in $D$ are included in the result dataset  $H_{\text{cluster}}$.
Although separating noise clusters and small clusters can be useful, it might be desirable to assign all wafer maps to clusters, either for performance comparisons 
or for practical considerations. For this purpose we propose the simple method of assigning the feature vectors in $H_{\text{rest}}$ and $H_{\text{small}}$ to clusters in $H_{\text{cluster}}$ 
by a nearest neighbor assignment. Note that the nearest neighbor search is done in the CNN feature vector space, and not in the PCA space, where components and therefore distances might change iteration by iteration. We consider this final assignment as optional, although it allows each wafer map to be assigned to a cluster 
while preventing the number of resulting clusters from becoming too large and having very few wafer maps in some clusters.

\subsubsection{Free Parameters of the full method}
In the presented ICH method, there are certain hyperparameters that can be adjusted depending on the data set and the use case.
In principle the CNN network architecture, dimensionality-reduction and clustering methods as well as the distance metric for silhouette scores can be exchanged, 
depending on the type of dataset. The parameters to be freely chosen for this method are the number of principal components allowed ($n_\text{PCA}$), the number 
of clusters allowed per iteration ($n_c$) and the minimum number above which a formed cluster is considered large enough ($n_{min}$). We motivate each selection 
for our experiments in \autoref{sec:method_setup}.

\subsection{Evaluation metric}
\label{sec:evaluation_metric}
To evaluate our results we use homogeneity as the performance metric.
As the term implies, homogeneity is a measure of the purity of the clusters formed. A cluster is considered pure if it contains only one defect class and no mixtures of different defect classes. The value for this metric $h$ is between 0 and 1 and is calculated by the formula
\begin{align}
h = 1 - \frac{H(C \rvert K)}{H(C)} \in [0,1]
\label{Formula: homogeneity}
\end{align}
while 
\begin{align}
H(C \rvert K) = - \sum_{c=1}^C \sum_{k=1}^K \frac{n_{c,k}}{n} \cdot \log \left( \frac{n_{c,k}}{n_k} \right)
\label{Formula: homogeneity Conditional Entropy}
\end{align}
and 
\begin{align}
H(C) = - \sum_{c=1}^C \frac{n_c}{n}\cdot \log \left( \frac{n_c}{n}\right)
\label{Formula: homogeneity Entropy Class}
\end{align}
with $H$ denoting the entropy, $C$ the set of classes and $K$ the set of clusters, $n$ the total number of samples, $n_c$ and $n_k$ the number of samples respectively belonging to class $c$ and cluster $k$ and finally $n_{c,k}$ the number of samples from class $c$ assigned to cluster $k$.
If an optimally pure clustering succeeds, in which only clusters with one class each are found, i.e. the entropy $H(C|K)$ is zero, the homogeneity approaches the value 1. The conditional entropy of the class distribution can be used to determine how close the proposed clustering is to the optimally homogeneous division of the data, i.e. each cluster consisting only of members of one class. The metric value becomes smaller the more members of different classes are found in the clusters.  $H(C|K)$ is maximal and equal to $H(C)$ "if the clustering does not provide any new information"~\cite{andrew_rosenberg_vmeasure_2007}. The homogeneity becomes $h=0$ in this case.
We used the implementation of the scikit-learn collection~\cite{scikit_learn}.
The homogeneity metric is used for evaluation in this work because it uses the labels as a basis. Since we are examining a labeled dataset, homogeneity is a valid measure of the purity of the true labels in the clustering result.

With the homogeneity described, our experiments are evaluated. 
Before presenting the results, the following section describes the specific configuration of our method and the datasets used for our experiments.
\section{Experimental setup}\label{sec:experimentalsetup}

\subsection{Method setup}
\label{sec:method_setup}
To perform our ICH method, a selection of methods for the three-step method is required and the associated parameters must be set.
We use the Xception architecture as a feature extractor, with weights pre-trained on the ImageNet dataset as available within Keras~\cite{chollet2015keras}.
Specifically, we use the outputs of the MaxPooling2D layer $block13\_pool$, which has been identified to work well in the two-step procedure analyzed by Gu\^{e}rin et al.~\cite{guerin_cnn_2018}.
For each wafer map with a resolution of $(299,299)$ and three channels an array with dimensions $(10,10,1024)$ is generated. The array is flattened to a feature vector with 102.400 dimensions.

As dimensionality reduction method we use PCA. In our experiments, we found that a number of $n_\text{PCA} = 20$ PCA dimensions is a reasonable choice. These first $20$ 
components cover about $80 \%$ of the variance in the full dataset.  

For the clustering method we use AC. We set the number of clusters within an individual iteration to $n_{c} = 15$. Allowing fewer clusters for ICH may result in mixed clusters with different true labels, once the unique clusters with characteristic salient properties are removed from the dataset to be clustered. Therefore, care must be taken to ensure that the number of allowed clusters for AC is large enough. If the value is chosen larger, small clusters can be found and the resulting number of result clusters may become larger, which is not the purpose of a clustering method.

To identify the cluster that is filtered out during an iteration, the cosine distance was 
used as the distance metric for the silhouette score.

As mentioned before, we filter $H_{\text{cluster}}$ for a minimum size of at least $n_{min} = 5$ wafer maps per cluster. In order to quote results for a full assignment of all 
wafer maps, the final nearest-neighbor assignment is conducted, both for the $H_{\text{rest}}$ and $H_{\text{small}}$ cluster sets.   

\subsection{Dataset description}
\label{subsec:Dataset}
\begin{figure} 	
	\centering
	\subfloat[\label{fig:Example_Dataset_Center}]{
		\includegraphics[width=0.1\columnwidth]{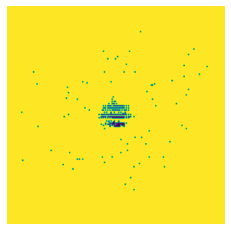}}
	\subfloat[\label{fig:Example_Dataset_Donut}]{
		\includegraphics[width=0.1\columnwidth]{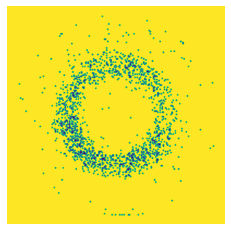}}
	\subfloat[\label{fig:Example_Dataset_EdgeLoc}]{
		\includegraphics[width=0.1\columnwidth]{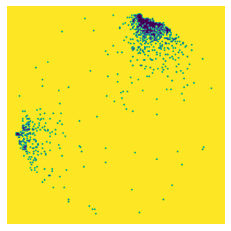}}
	\subfloat[\label{fig:Example_Dataset_Loc}]{
		\includegraphics[width=0.1\columnwidth]{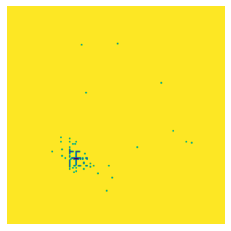}}
	\subfloat[\label{fig:Example_Dataset_Nearfull}]{
		\includegraphics[width=0.1\columnwidth]{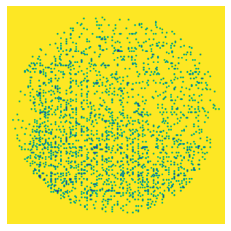}}
	\subfloat[\label{fig:Example_Dataset_Random}]{
		\includegraphics[width=0.1\columnwidth]{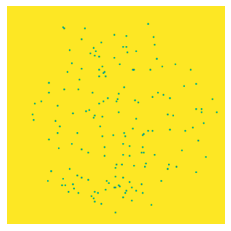}}
	\subfloat[\label{fig:Example_Dataset_Ring}]{
		\includegraphics[width=0.1\columnwidth]{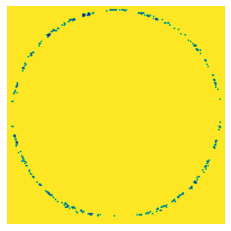}}
	\subfloat[\label{fig:Example_Dataset_Scratch}]{
		\includegraphics[width=0.1\columnwidth]{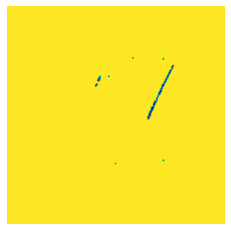}}
	
	\caption{Exemplary images of wafer maps in the dataset WM1K with (a) Center, (b) Donut, (c) Edge-Loc, (d) Loc, (e) Near-Full, (f) Random, (g) Ring and (h) Scratch defect pattern.}
	\label{fig:dataset_examples}
\end{figure}
\begin{table}
	\begin{center}
		\caption{Characteristics of used wafer map datasets with defect-type distribution}
		\label{fig:dataset_distribution}
		\begin{tabular}{| c | c | c |}
			\hline
			Dataset & \textbf{WM1K} & \textbf{WM811K\_sub} \\ 
			\hline
			\textbf{Resolution} & $200\,\times\,200$ & $27\,\times\,27\,\pm\,2$\\
			\hline
			\textbf{Total number} & 1302 & 923\\
			\hline
			\hline
			Center & 166 & 150\\ 
			\hline
			Donut & 208 & 9 \\ 
			\hline
			Edge-Loc& 208 & 150 \\
			\hline
			Loc & 120 & 150 \\ 
			\hline
			Near-Full & 150 & 53\\ 
			\hline
			Random & 150& 141 \\ 
			\hline
			Ring & 150 & 120\\ 
			\hline
		    Scratch & 150 & 150\\ 
		    \hline
		\end{tabular}
	\end{center}
\end{table}
The proposed clustering method is evaluated using two wafer map datasets, which are sharing the same logic for the defect classes.
The main evaluation in this paper is shown by using a real-world wafer map dataset WM1K provided by a semiconductor manufacturing company. WM1K contains 1302 wafer maps belonging to the following eight defect patterns: Center, Donut, Edge-Loc, Loc, Near-Full, Random, Ring or Scratch. Examples of these defect patterns are shown in 
\autoref{fig:dataset_examples}. The labels were assigned manually by domain experts. The variance of the appearances of defect distributions may vary for different defect patterns. For example, the visual appearance of the defect distribution for Center, Donut, and Near-Full classes are quite uniform due to their geometry and class definition. A Donut can vary in diameter and thickness, but the characteristic shape stays the same. On the other hand, the Edge-Loc and Scratch defects in particular can occur at very different locations with various orientations on the wafer maps. Ring defects can be putatively evaluated as a simple class. However, the Ring can also only occur as a circular segment. Near-Full and Random defects vary in density of defects. Even in areas where two wafer maps of these types are both affected by defects, the distribution of defects appears randomly, so the pixel-based variations may be large.
These differences will have a significant impact on clustering performance, as discussed later in \autoref{sec:results}.
The number of wafer maps in WM1K varies between $120$ and just over $208$ for all defect classes, but no class is severely underrepresented, which can be seen in \autoref{fig:dataset_distribution}.
The wafer maps have a resolution of $(200,200)$ pixels. To match the input resolution of the pre-trained CNN the wafer maps are resized to $(299,299)$ and copied along a newly introduced third dimension. Besides rescaling, no further preprocessing is applied to WM1K.
The method is evaluated on the not-preprocessed wafer maps, as the aim of this paper is to investigate to what extent defect patterns can be clustered, partly containing many randomly occurring defects that do not belong to the raw, typical signature of the defect class.

The second dataset used for evaluating the method is a subset of the public dataset WM811K~\cite{Yu_similarity_search_large_datasets}.
WM811K contains a wide variety of different image resolutions.
Furthermore, classes are heavily imbalanced for many resolutions.
Therefore, we carry out the major part of our analysis on the dataset WM1K, which is much better maintained and allows direct statements based on a real-world use case.
To prove that our method is generally applicable, we carry out an additional evaluation on the public WM811K dataset.
We extract images with a resolution around $(27, 27)$.
To mitigate the current class imbalance, we 
consider up to $150$ images per defect pattern (see \autoref{fig:dataset_distribution}). We denote this subset of the public dataset as WM811K\_sub.
The same pre-processing as before is performed to match the standard CNN input resolution.

\section{Results}
\label{sec:results}
This section presents the performance of our proposed cluster harvesting method mainly on the dataset WM1K as described in \autoref{subsec:Dataset}.
First we report the overall performance of our ICH approach and show the improvement compared with one-time clustering with the same algorithm sequence. \\
To give further insight into the procedure, we show the workings on the first iteration in detail 
and follow the clustering progress through to the final iteration.  
We also present examples of the clusters filtered out by our additional filter on the cluster size, to illustrate our reasoning behind this choice. To illustrate  
the limitations of performance on the given dataset, we show a detailed view on the types of clusters which still show strong mixing of true defect types.\\ 
To further demonstrate the competitiveness of our iterative method, we also show a benchmark comparison by total homogeneity with other one-time clustering approaches. 
We validate that the performance improvement from our iterative approach is stable with regards to the choice of method for each part of the three-step sequence, 
by presenting an ablation study, where we quote results with a viable alternative algorithm chosen for each part. \\
Finally, we show how our proposed harvesting method improves clustering performance on the public dataset WM811K\_sub as well.
\subsection{Results of iterative cluster harvesting}
\begin{figure}
	\centering	
	\subfloat[One-Time Clustering (OTC)]{
		\includegraphics[width=\columnwidth]{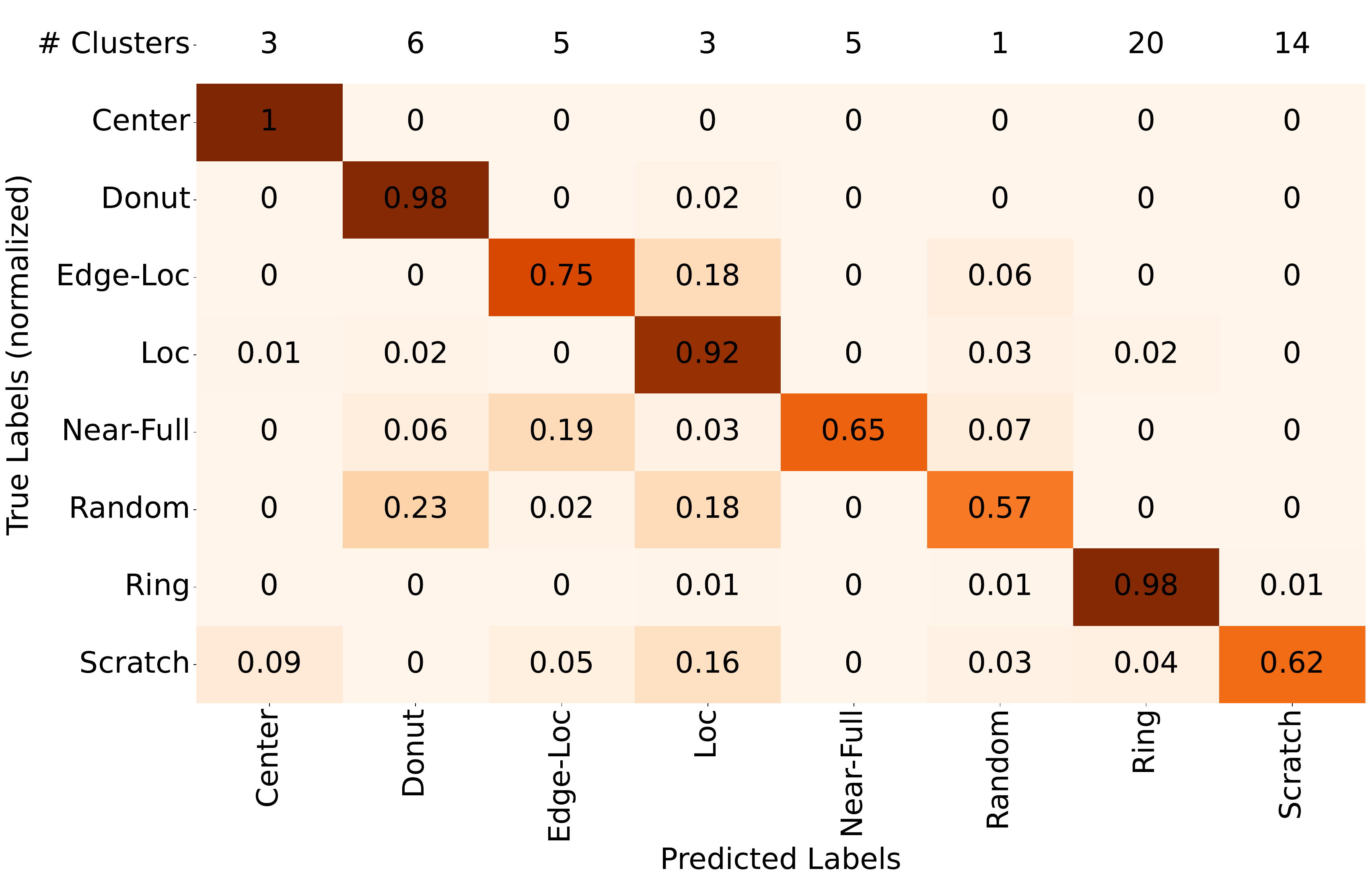}\label{fig:res_confusionmatrix_onetime}}
	
	\subfloat[Iterative Cluster Harvesting (ICH)]{
		\includegraphics[width=\columnwidth]{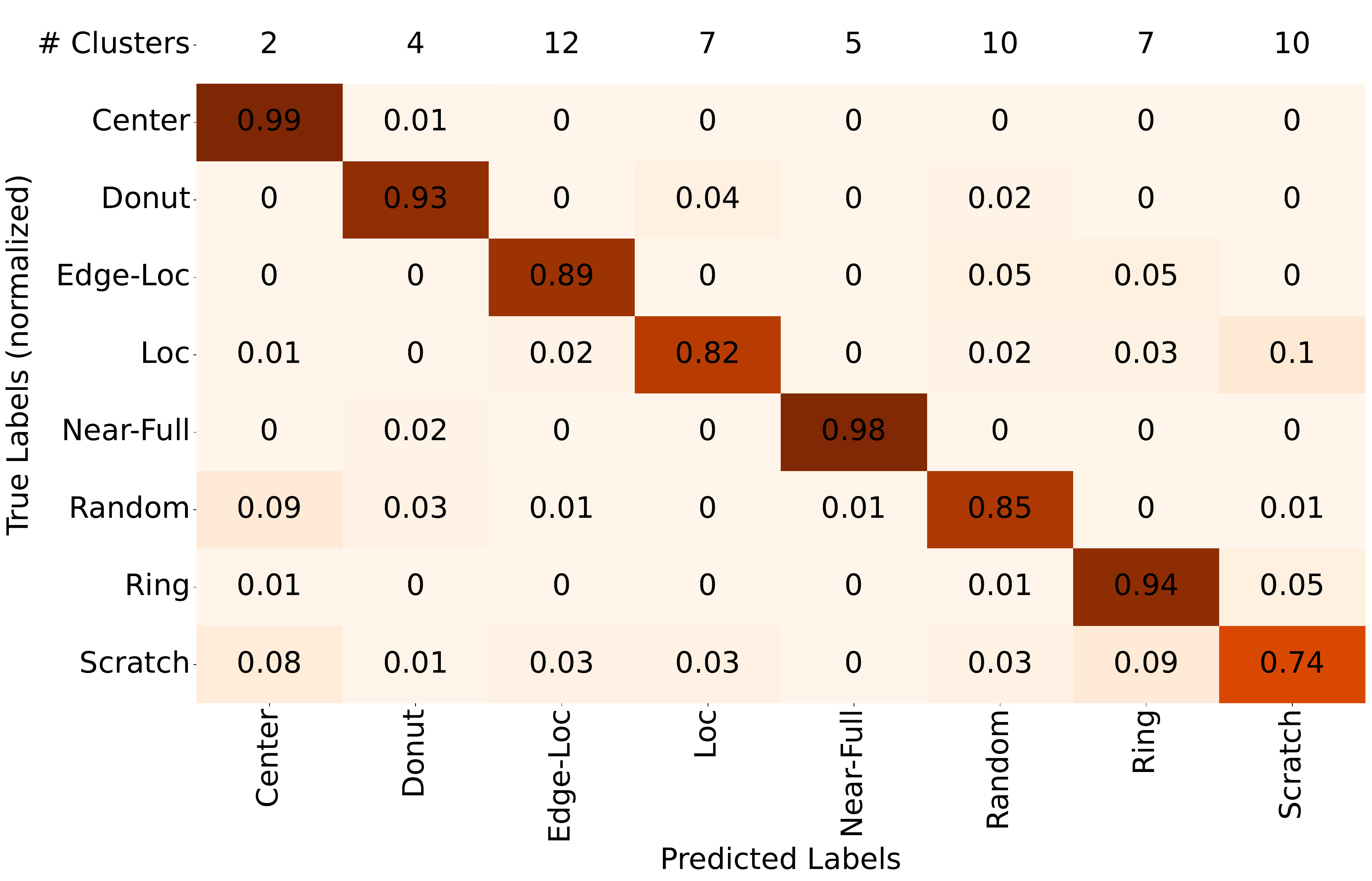}\label{fig:res_confusionmatrix_iterative}}
	\caption{Confusion matrices of OTC (a) and ICH (b) approach with full assignment of WM1K dataset. For each cluster, the predominant true label is evaluated and assigned as the predicted label. If there are other true labels in a predominant cluster of a particular true label, this can be read along the vertical column of the true label. Values are normalized to the number of data points of the true labels. Since there are usually multiple clusters for the same label, they are combined in order to calculate the confusion matrix. The number of individual clusters per true label is shown in the top row of the respective confusion matrix.}
	\label{fig:res_confusionmatrix_one_time_iterative}
\end{figure}

Applying our method described in~\autoref{sec:methodology} to the WM1K wafer map dataset automatically determines the number of clusters, yielding $57$ clusters. 
The homogeneity for this clustering result is $h=0.90$ and with optional full assignment $h=0.86$.\\
As a first, and most instructive comparison with the proposed method, we apply the three-step procedure only once (OTC). 
Our proposed ICH approach shows significantly improved performance compared with the OTC approach on the WM1K dataset both for partial and full assignment. 
This comparison, together with the other performance comparisons described in the following are summarized in \autoref{tab:res_benchmark_ac}.
With OTC on WM1K (full assignment) some defect classes are well clustered and some are mixed with others. This method yields a homogeneity metric value of $h\,=\,0.77$ (full assignment) and $h\,=\,0.82$ (partial assignment). This already shows the utility of the filtering introduced by ICH, as it identifies the subset of the dataset 
that is easier to cluster and removes many outliers and small aggregations of wafer maps which are otherwise wrongly assigned in OTC.\\
To understand the class-wise improvement of performance we present the confusion matrix, normalized by true labels in 
\autoref{fig:res_confusionmatrix_one_time_iterative}, 
where the predicted labels are taken to be the majority class true label within a given cluster.  
For the defect patterns Center, Donut, Loc, and Ring, OTC results in a percentage of correctly clustered wafer maps range from $92\,\%$ to $100\,\%$ (see \autoref{fig:res_confusionmatrix_onetime}).
For the other defect patterns (Edge-Loc, Near-Full, Random, and Scratch), the majority class makes up for $57\,\%\,-\,75\,\%$ of the cluster. 
These defect patterns could not be distinguished well from each other by the OTC method, resulting in clusters that contain wafer maps with different 
defect patterns. For example, Edge-Loc and Loc is mixed, Near-Full is mixed with Edge-Loc, Random is mixed with Donut and Scratch is mainly mixed with Loc.
The number of clusters formed for each defect pattern is listed at the top of the confusion matrix in \autoref{fig:res_confusionmatrix_onetime}.
In particular, many clusters were formed for Ring and Scratch (20 and 14, respectively) and only one cluster for defect pattern Random.\\
To evaluate the change in performance for the individual defect patterns for OTC and ICH, we consider the diagonal entries of the confusion matrices in \autoref{fig:res_confusionmatrix_one_time_iterative}. 
Our proposed harvesting procedure improves the diagonal percentages 
considerably for the four defect patterns that had low values with OTC procedure.
Comparatively small decreases 
are seen for the four defect patterns which had very high values with the OTC procedure. This overall reduction of off-diagonal entries as well as the changes in 
the number of clusters per majority class, illustrates how the iterative harvesting of individual, well-separated clusters prevents lumping many different true labels 
together, sometimes at the expense of creating more, smaller clusters.

\subsection{Illustration of iterative cluster harvesting process}

\begin{figure}[ht]
	\includegraphics[width=\columnwidth]{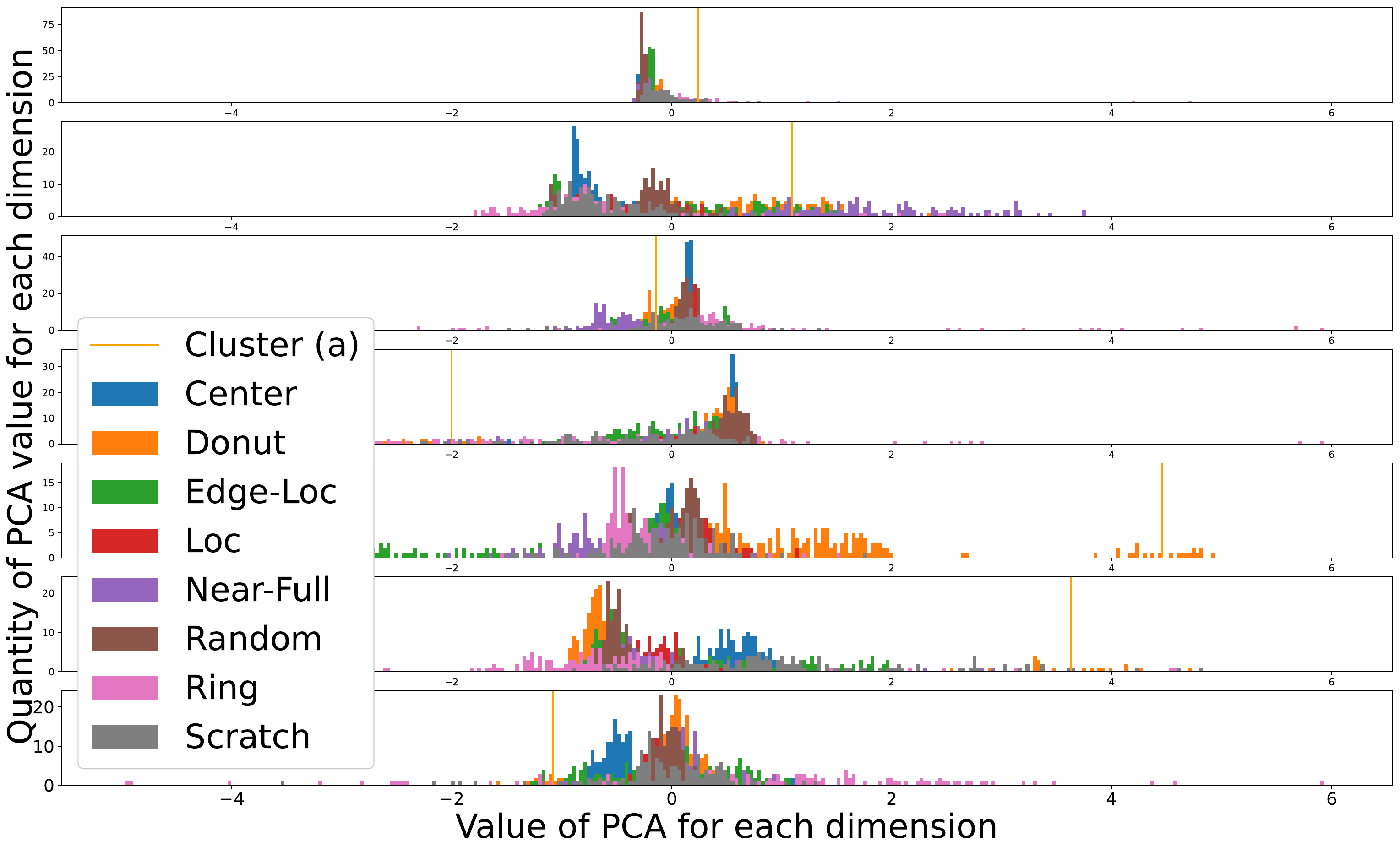}
	\caption{Histogram over the feature values of the first seven principal components after applying the OTC to WM1K, which is equal to the first iteration of the ICH procedure. Colors are assigned according to the true labels. Mean values of an exemplary cluster (a) shown in~\autoref{fig:Good_clusters_Iterative_Donut} are marked in each principal component.}
	\label{fig:resOneTimePCAHist}
\end{figure}

To demonstrate the way of harvesting, the PCA space of the first iteration is presented in \autoref{fig:resOneTimePCAHist}. It shows the seven leading PCA components, color-coded by the true label.

\begin{wrapfigure}{l}{0.42\columnwidth}
	\subfloat[\label{fig:Good_clusters_Iterative_Donut}]{
		\includegraphics[width=0.2\columnwidth]{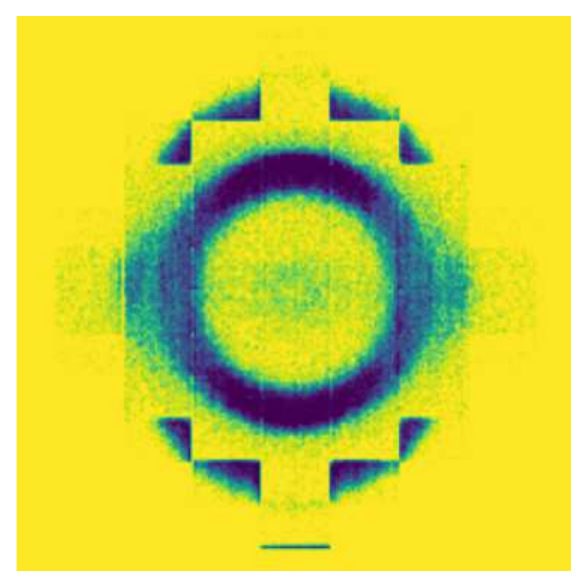}}
	\subfloat[\label{fig:Good_clusters_Iterative_Scratch}]{
		\includegraphics[width=0.2\columnwidth]{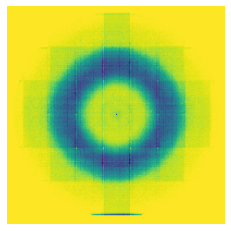}}
	\label{fig:res_good_and_bad_clusters_Iterative}
	\caption{(a): Mean image of the Donut cluster identified in the 1st iteration by ICH for WM1K and (b): mean image of all wafer maps of WM1K from the Donut class except for the wafer maps from the identified cluster (a).}
\end{wrapfigure}

Although this illustration shows only projections into the first several PCA dimensions (not the silhouette score actually used to measure the separation between clusters), it already becomes apparent how some groupings of very well separated wafer maps can be identified. Looking at the color-coding of these groupings, we find large, well-separated groups for the classes we expect to have a very characteristic appearance from the discussion in \autoref{subsec:Dataset}.
Other classes with varying appearance such as Loc or Random largely share a common range of values. Some clusters are well separated from the remaining samples.
For the cluster with the highest silhouette score after performing the first iteration of our ICH procedure, we have marked the mean values within the different PCA dimensions.
While its individual sample values are still dispersed, its distance to the neighboring samples is especially large in the principal components $4$, $5$, and $6$.
This visually distinct cluster contains $22$ Donut wafer maps. The mean image of this cluster is shown in~\autoref{fig:Good_clusters_Iterative_Donut}.
Note that the mean images are calculated by summing all images in the respective cluster 
\noindent 
of wafer maps pixel by pixel and dividing by the number of images in the cluster. Visually it is apparent why this cluster is well separated, as there is a significant difference from the mean image of the Donut class without these clustered wafer maps, as the stepped edges around the wafer maps only appear in the wafer maps in~\autoref{fig:Good_clusters_Iterative_Donut}.

\begin{figure}[th]
	\includegraphics[width=\columnwidth]{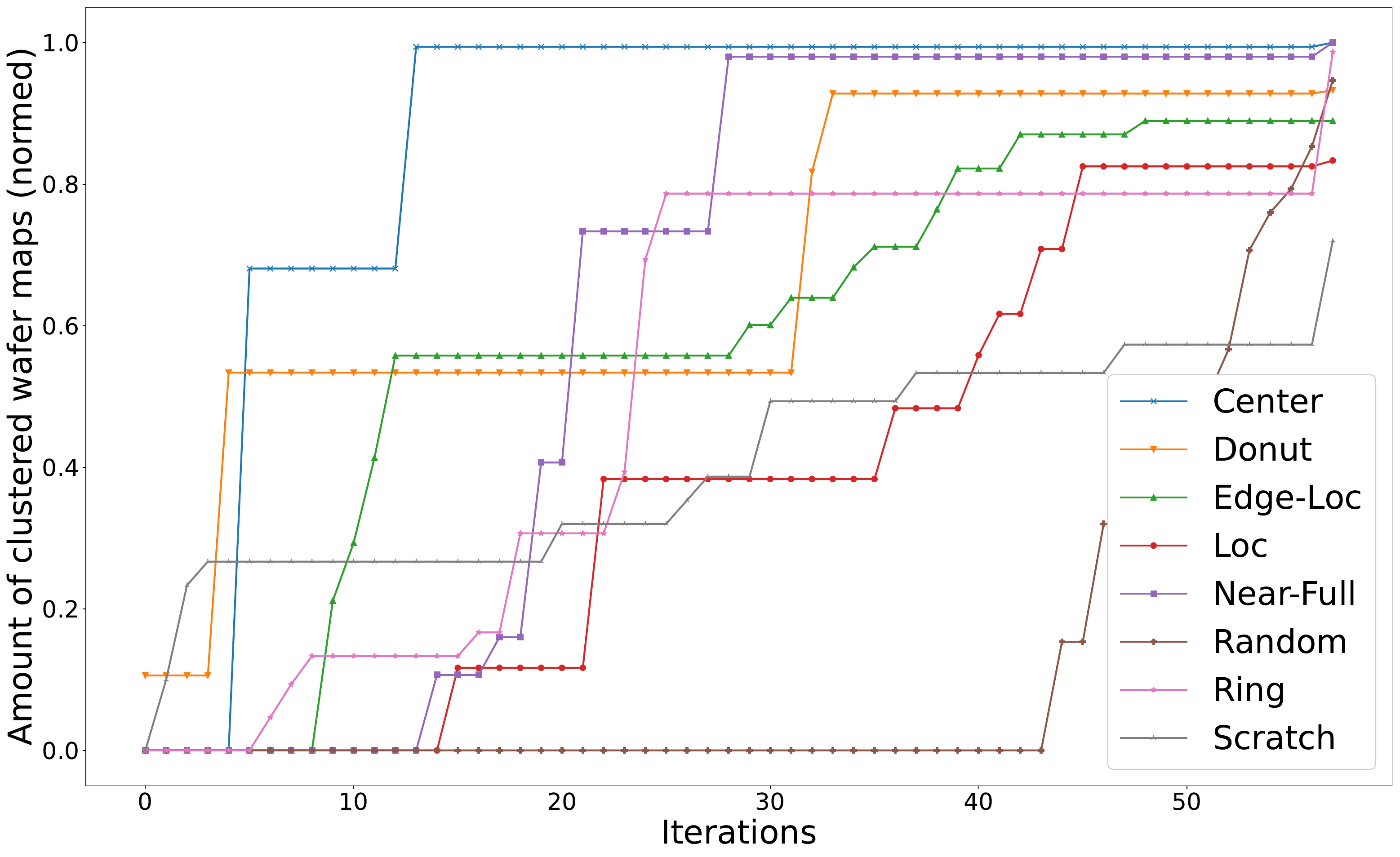}
	\caption{Number of clustered wafer maps (normalized to 1) as a function of iterations for each defect class for ICH on the WM1K dataset. Per iteration, the dominating true label for the resulting cluster is determined and counted. The wrongly assigned wafer maps in the cluster are not counted. The last iteration relates to the nearest neighbor assignment, which is performed for the wafer maps in $H_{\text{rest}}$ and $H_{\text{small}}$.}
	\label{fig:resIterativeAmount}
\end{figure}
After the first iteration the wafer maps in this cluster are extracted to $H_{\text{cluster}}$, which leads to an initial assignment of about $10\%$ of wafer maps from the Donut class.
The consecutive accumulation of correctly assigned wafer maps per defect class is shown in \autoref{fig:resIterativeAmount}.
Along the $x$-axis of \autoref{fig:resIterativeAmount}, one can follow which majority class is being harvested in the cluster for each iteration, 
while small clusters are not considered and are assigned in the last iteration at the end in \autoref{fig:resIterativeAmount}. 
It can be seen that often there are consecutive iterations where the majority class of the harvested clusters is the same, 
suggesting that the PCA evolves slowly in these cases, keeping focus on the features relevant for this class. 
Switches of majority defect class (i.e. one line flattening out and another rising) denote iterations where the variance in the dataset is then dominated by 
another class, potentially causing larger changes in the PCA dimensions. By this logic a loose ranking by variance of the different 
classes can be extracted from this figure. 
Defect patterns with a rather uniform visual appearance such as Center or Donut form clusters which exhibit high silhouette scores 
and are extracted from $D$ during early iterations. Defect patterns with varying appearance such as Scratch or Random are clustered at later iterations. 
Looking at the second-to-last iteration in \autoref{fig:resIterativeAmount} one sees the total, correctly assigned fraction of wafer maps from each class 
after ICH but before the final nearest-neighbor assignment. 
Before nearest-neighbor assignment almost all defect classes have a fraction of $\gtrapprox 0.75$ of their wafer maps assigned 
correctly, with the Scratch class being the only exception with a fraction of about $0.6$ during the regular ICH procedure.\\
\autoref{fig:resIterativeAmount} further shows the performance of the final assignment of our approach, where the remaining samples ($H_{\text{rest}}$ and 
$H_{\text{small}}$) are assigned to their nearest neighbors of already assigned samples in the CNN feature space. 
After nearest neighbor assignment, all correctly assigned fractions increase, showing the validity of the simple approach. It is also interesting to note 
that especially the two classes with the lowest fractions, Scratch and Ring, rise dramatically in the final assignment.
This suggests that these maps lie close-by in feature space to well-separated clusters with this majority class, but many samples are scattered too widely to be included in well-separated 
clusters during the main ICH procedure.

\subsection{Examples of wafer maps from $H_{small}$}
\setlength\parfillskip{0pt}\par\setlength\parfillskip{1pt plus 1fil}
\begin{wrapfigure}{l}{0.5\columnwidth}	
	\subfloat[\label{fig:Aside_Ring_1}]{
		\includegraphics[width=0.1\columnwidth]{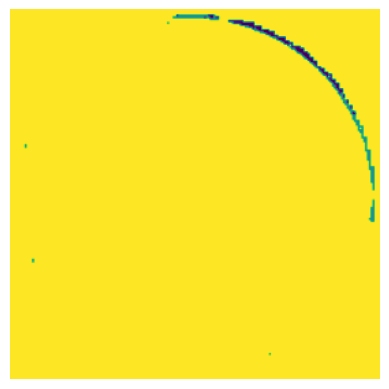}}
	\subfloat[\label{fig:Aside_Ring_2}]{
		\includegraphics[width=0.1\columnwidth]{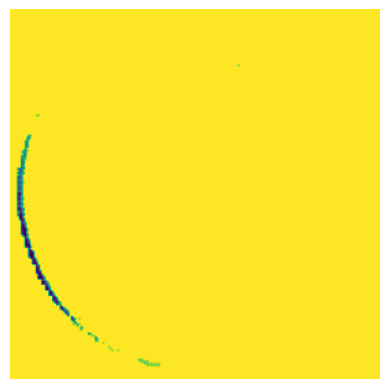}}
	\subfloat[\label{fig:Aside_Scratch_1}]{
		\includegraphics[width=0.1\columnwidth]{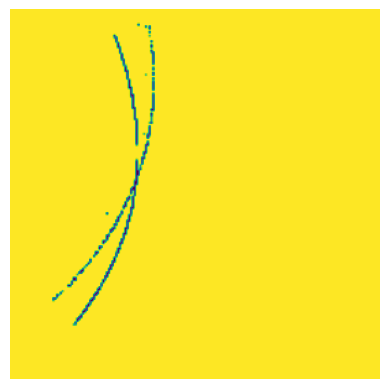}}
	\subfloat[\label{fig:Aside_Scratch_3}]{
		\includegraphics[width=0.1\columnwidth]{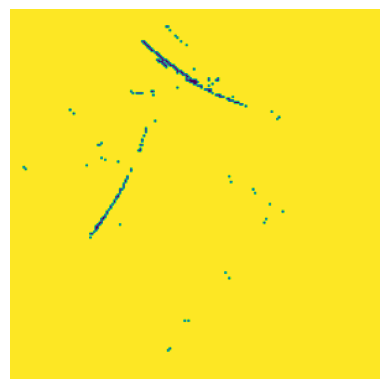}}
	
	\caption{Exemplary images of wafer maps identified by ICH with high purity and small number in WM1K. Images (a-b) show Ring and (c-d) Scratch wafer maps.}
	\label{fig:res_examples_aside}
\end{wrapfigure}
\noindent
To illustrate what types of wafer maps we remove during iterations which formed a small cluster (less than $5$ wafer maps) we show some examples in \autoref{fig:res_examples_aside}. As can be seen in this figure, these wafer maps display very distinct patterns and often near 
zero pixel-wise variance inside the small clusters. While potentially interesting to diagnose small aberrations in the wafer manufacturing, they are less 
valuable than the larger clusters found in other iterations, which point to production problems that affect a larger population of wafers potentially over 
extended time periods. 

\subsection{Examples of clusters with low homogeneity}
\begin{figure}
	\begin{center}
		\begin{tabular}{ |c | c | c | }
			\hline
			\textbf{Cluster}& \textbf{Dominating TL} & \textbf{Mixed TL}   \\ \hline
			\multirow{2}{*}{A} &
			\includegraphics[width=0.1\columnwidth]{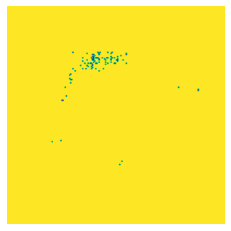} & \includegraphics[width=0.1\columnwidth]{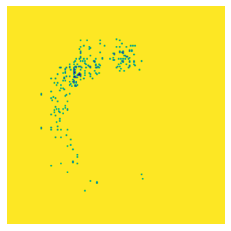}  \\ 
			& Loc & Donut \\  \hline
			\multirow{2}{*}{B} & 	\includegraphics[width=0.1\columnwidth]{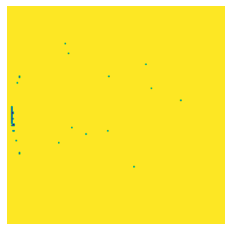} & \includegraphics[width=0.1\columnwidth]{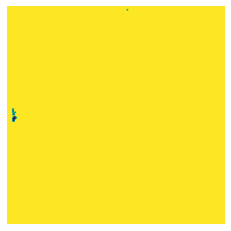}  \\ 
			& Ring   & Edge-Loc \\  \hline
			\multirow{2}{*}{C} & \includegraphics[width=0.1\columnwidth]{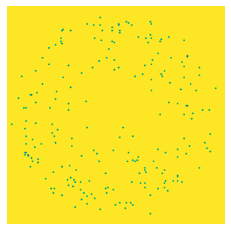}  & \includegraphics[width=0.1\columnwidth]{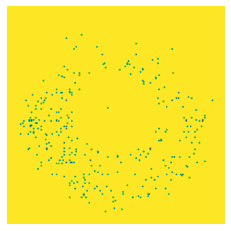}   \\ 
			&  Random & Donut \\  \hline
		\end{tabular}
	\end{center}
	\caption{Exemplary wafer maps of clusters with mixed true labels (TL) after applying ICH to the WM1K dataset.}
	\label{fig:res_mixed_clusters}
\end{figure}
As a result of the ICH process, many clusters have been created that are strongly dominated by one defect class (high homogeneity).
However, even with this ICH approach some mixed clusters remain.
A visual assessment as shown in \autoref{fig:res_mixed_clusters} gives a deeper insight into the cause for the wrong assignments: some samples share a similar visual appearance, even though they are assigned with different labels.
For example, cluster A in \autoref{fig:res_mixed_clusters} is composed of Loc defect patterns located in the upper left wafer area. The Donut defect pattern mixed into this Loc cluster looks very similar, with defects mainly present in the upper left wafer area as well. A highly similar visual appearance although with different true labels is further shown for the exemplary clusters B and C.

\subsection{Comparison with other one-time clustering methods}

\begin{table}[ht]
	\begin{center}
		\caption{Benchmark of our proposed ICH approach on WM1K dataset with $57$ clusters. The homogeneity values are reported without assignment of wafer maps in $H_{\text{rest}}$ and $H_{\text{small}}$ (partial assignment) and with assignment of all wafer maps (full assignment).}
		\begin{tabular}{| c | c | c |}
			\hline
			Method &Partial assignment& Full assignment\\ 
			\hline
			OTC & 0.82 &  0.77  \\ 
			\hline
			CNN + AC \cite{guerin_cnn_2018} & 0.87 & 0.78  \\
			\hline
			PCA + AC &  0.73  &  0.74\\ 
			\hline
			Var-AE + AC \cite{tulala_unsupervised_2018} & 0.72 & 0.73  \\ 
			\hline
			ICH (ours)& \textbf{0.90} & \textbf{0.86}  \\ 
			\hline
		\end{tabular}
		\label{tab:res_benchmark_ac}
	\end{center}
\end{table}

To illustrate the competitiveness of ICH, we also quote homogeneity numbers for some other relevant one-time clustering methods in \autoref{tab:res_benchmark_ac}. 
To enable a direct comparison, we use the same number of clusters as the number of clusters identified during our ICH approach for the benchmarking methods.
The choice of hyperparameters (e.g., CNN network architecture, PCA dimensions) remained the same for each component.

Applying the method presented by Gu\^{e}rin et al. \cite{guerin_cnn_2018} (CNN+AC) on the full dataset WM1K (full assignment), we see that the homogeneity decreases to $h=0.78$  compared to $h=0.86$ (ICH). The performance of this two-step procedure is almost equal to the one of the OTC (CNN+PCA+AC), which yields a homogeneity of $h=0.77$.
If PCA is applied directly to the images without feature extraction by a CNN, the homogeneity of the clustering result decreases to $h=0.74$. Similar to Tulala et al.~\cite{tulala_unsupervised_2018}, we also train a variational autoencoder on the wafer map dataset. It is trained for $40$ epochs and the latent space with $20$ dimensions is clustered using AC. This results in a smaller homogeneity of $h=0.73$. 

The same applies for the results of the part of wafer maps that were divided into sufficiently large clusters by ICH (partial assignment). The clustering result of ICH has the highest homogeneity with $h=0.86$, while CNN+AC performing better than OTC.

The proposed cluster harvesting procedure ICH exhibits a competitive performance when compared to other methods which are similar in terms of setup effort and computational complexity.

\subsection{Ablation Study}
\begin{table}[htbp]
	\caption{Ablation study: relative improvement $\Delta_{\text{rel}}$ of homogeneity metric (ICH vs OTC) for different choices of feature extractors (Feat. Extr.), dimensional reduction (dim. red.) and clustering methods for WM1K with full assignment.}
	\label{tab:res_benchmark_relative}
	\begin{center}
		\begin{tabular}{|m{1.2cm}|c|c|c|c|}
			\hline
			\textbf{Ablation Factor }& \textbf{Exemplary Pipeline} & OTC &ICH & \textbf{$\Delta_{\text{rel}}$}   \\ \hline
			Proposed Pipeline & Xception + PCA + AC &0.77& 0.86 & 12\,\% \\ \hline %
			\multirow{2}{*}{Feat. Extr.}     & ResNet + PCA + AC & 0.75 & 0.82 & 9\,\%  \\ \cline{2-5}
			& VGG16 + PCA + AC  & 0.74 & 0.80 &  8\,\% \\  \hline
			\multirow{2}{*}{Dim. Red.} &  Xception + ISOMAP + AC &0.78&  0.81 & 4\,\% \\ \cline{2-5}
			&  Xception + SVD + AC & 0.60 & 0.68 & 13\,\%  \\ \hline
			\multirow{2}{*}{Clustering}    &  Xception + PCA + GMM &  0.74 & 0.80 & 8\,\% \\ \cline{2-5}
			&  Xception + PCA + k-means  & 0.73 & 0.80 & 10\,\% \\ \hline %
		\end{tabular}
	\end{center}
\end{table}
In order to investigate whether the ICH process provides an improvement in homogeneity independent of the choice of the underlying network architecture, 
the dimensionality reduction and the clustering procedure, an ablation study was conducted on the full WM1K dataset using the nearest neighbor assignment. 
One component of the three-step process is changed at a time and the comparison of homogeneity between the OTC and ICH execution of the process is evaluated (see \autoref{tab:res_benchmark_relative}).

When the network architecture is changed, the relative improvement due to the iterations ranges from $8\,\%$ to $9\,\%$.
If one exchanges the dimensionality reduction to either ISOMAP or SVD, homogeneity is also improved by applying our proposed ICH approach. 
When changing the clustering method to GMM an improvement by the iterative procedure of $8\,\%$ and with k-means of $10\,\%$ is seen, 
compared with the $12\,\%$ improvement when applying ICH with AC clustering.
\subsection{Results of ICH on WM811K\_sub}
Applying our method described in~\autoref{sec:methodology} to the WM811K\_sub wafer map dataset automatically determines the number of clusters, yielding $66$ clusters. 
The homogeneity for this clustering result is $h=0.80$ for the clusters in $H_{\text{cluster}}$ of ICH (see
also \autoref{tab:res_benchmark_ac_nn_wm811k}).
The application of OTC leads to a homogeneity of $h\,=\,0.75$. \\
Therefore our proposed ICH approach significantly improves performance over the OTC approach on the dataset of clusters selected by ICH 
($6\,\%$ relative improvement). ICH outperformed the other benchmark methods as well.
Using the optional variant with full assignment and forcing all data points from small filtered out clusters
to a cluster via nearest neighbor assignment, the homogeneity for the WM811K\_sub dataset drops to
$h\,=\,0.62$. 
For this dataset, about 20\% of the wafer maps end up in $H_{\text{small}}$ and the
assignment via nearest neighbor search leads to a sharp drop in cluster performance. 
Therefore it is not surprising that on the entire dataset other methods lead to a slightly higher homogeneity. OTC achieves $h\,=\,0.67$  
and CNN+AC performs best with $h\,=\,0.69$.\\
We note however that omitting the difficult wafer maps, identified by our method in $H_{\text{rest}}$ and $H_{\text{small}}$, the clustering 
performance of all methods improves noticeably. For example, comparing the homogeneity values of the OTC with the data from $H_{\text{cluster}}$ 
and the full assignment, the homogeneity improves strongly from $h\,=\,0.67$ to $h\,=\,0.75$. Thus, the utility of our method as a filter to create 
a well-clusterable sub-dataset can also be clearly shown when applied to the WM811K\_sub dataset.

\begin{table}
	\begin{center}
		\caption{Benchmark of our proposed ICH approach on WM811K\_sub with $66$ clusters. The homogeneity values are reported without assignment of wafer maps in $H_{\text{rest}}$ and $H_{\text{small}}$ (partial assignment) and with assignment of all wafer maps (full assignment)..}
		\label{tab:res_benchmark_ac_nn_wm811k}
		\begin{tabular}{| c |c | c |}
			\hline
			Method & Partial assignment & Full assignment \\ 
			
			\hline
			OTC &0.75 &0.67 \\ 
			\hline
			CNN + AC \cite{guerin_cnn_2018} &0.77 & \textbf{0.69} \\
			\hline
			PCA + AC &  0.68 & 0.64 \\ 
			\hline
			Var-AE + AC \cite{tulala_unsupervised_2018} & 0.56 & 0.52 \\ 
			\hline
			ICH (ours) &\textbf{0.80} & 0.62 \\ 
			\hline
		\end{tabular}
	\end{center}
\end{table}

\section{Discussion}
\label{sec:discussion}

We have shown that our method of iteratively performing variance-based dimensionality reduction and removing sets of wafer maps, that are well-separated, leads to improvements in homogeneity of clustering results in comparison with competitive non-iterative approaches. 
We note however that for the Var-AE the comparison on our dataset might not be fully conclusive, as we would expect an improved performance of the auto-encoder 
with a larger dataset size. Nonetheless we still claim an advantage in simplicity and training effort of our method in comparison to Var-AE or similar methods. \\ 
We have also shown that these benefits persist also when replacing individual components of the preferred 
pipeline of methods. After applying a modular feature extraction method, which is appropriate for images, our method can be applied with competitive speed and performance 
to the clustering of image datasets, again with freedom of choice for the clustering method without suffering usual difficulties from dealing with high-dimensional 
image data. Our iterative method does not introduce new free hyperparameters, apart from the stopping criterion for the minimal number of samples and the optional filter 
criterion on the size of the selected clusters. We note however that the number of PCA dimensions as well as the expected number of clusters for AC, which are 
also parameters of OTC, have a strong influence on the performance of our method and their meaning is slightly changed (e.g. number of clusters per AC iteration does 
not define the total number of clusters found by ICH). 

As an extension of the method, refinement of the collected clusters based on size provides a set of difficult, or "noise" wafer maps.
In clustering, such data points lead to the formation of many small clusters and with a fixed number of resulting clusters, as is common
in clustering approaches, fewer clusters are available for the remaining data points.
Therefore, we consider the automated identification of difficult samples as an advantage of our approach, but provide a simple method to assign these wafer maps 
automatically at the end of the core method to achieve a full assignment. 
Instead of executing this final assignment, the filtered sub-dataset can be used for manual review of these difficult wafer maps.

We showed the benefit of using our method as a filter, as we obtain much better homogeneity when applying some of the other considered clustering methods 
(namely OTC and CNN+AC) on our filtered dataset compared with the full dataset. 
For the other benchmark methods (PCA+AC and Var-AE+AC), there is little impact on cluster performance whether one clusters with or without the difficult 
wafer maps. It seems that direct PCA on image vectors as well as Var-AE training are robust against removal of these small sets of difficult wafer maps. \\
Although overall performance of ICH is very competitive, we still observe some strongly mixed clusters in our main dataset under study (WM1K).
We have shown examples of these cases to illustrate that we believe these cases to be difficult to judge even for humans, without additional information about the 
root-cause of the defect patterns. Thus our method could even be used in these cases to trigger manual review of dataset labels and to identify overlaps or borderline cases.\\
For the WM811K\_sub dataset, the homogeneity value with full assignment is lower for ICH than for the other methods studied (with exception of Var-AE). 
We showed in our results that the main cause for this are a large number of small clusters, which are formed during ICH and then not assigned by the core method 
but by nearest-neighbor assignment. This is not surprising, but rather indicates that this subset from the public dataset WM811K contains many wafer maps that are difficult to cluster. As with the WM1K dataset, filtering out the difficult wafer maps in WM811K\_sub improves clustering performance in the benchmark methods.
Nonetheless, we recommend examining the filtered-out data and adjusting the minimum value of an allowed cluster size in cases where an excessive amount of 
data points is filtered out during the application of ICH.

In summary, we conclude that our iterative method is applicable for a wide range of feature extraction and clustering methods.  Apart from configuring the hyperparameters, 
we did not make any specific adaptions to the algorithm in order to improve its performance on wafer map images. So, we expect it to be easily adaptable to other 
use cases and data types. \\
Considering our original use case of wafer map defect patterns, our method is well suited when combined with the filtering options discussed above.
In combination with example visualizations as shown in our paper, manual and interactive evaluation of the results should be possible. In this way, we expect it to 
be of great use in supporting defect engineers in detecting both rare and recurring defect patterns and providing them a narrowly focused starting point for root-cause analysis.\\
Finally, we expect that the proposed ICH method could be used in conjunction with or as a stand-in for supervised image classification methods. 
While by itself it would perform slightly worse in terms of clustering homogeneity, it would be useful for example to reduce the manual effort for labeling. 
Our method could also play a valuable part when a trade-off between performance and training effort can be considered, especially for datasets where the required 
domain knowledge is scarce.

\section{Conclusion}
\label{sec:conclusion}

Our proposed ICH approach improves the performance of unsupervised clustering wafer maps strongly. 
This performance improvement results from the potentially easier clustering of small variances in the data set in the course of the
iterative application of dimensionality reduction to the successively reduced dataset. 
In addition, our method helps to identify data points in a dataset that mainly lead to a worse clustering result because they are very
different from the rest of the dataset and consist of only a small number of wafer maps.

This method is very well suited to support manual labeling or review of existing labels. The method was not specialized for the wafer map use case, allowing it to be easily transferred to use cases with other image data. 
No manual effort is required for the use of this method, except for the selection of hyperparameters.
A separate investigation could examine how this ICH method works with high-resolution images, i.e. when the resolution is larger than the required input dimension of the feature extractor. This is relevant, for example, for clustering scanning electron microscope (SEM) images of wafers, which are typically acquired at higher resolutions.
\appendix
\section*{Author statement}
We describe the individual contributions of Alina Pleli (AP), Simon Baeuerle (SB), Michel Janus (MJ), Jonas Barth (JB), Ralf Mikut (RM) and Hendrik P. A. Lensch (HL) using CRediT~\cite{brand_beyond_2015}: \textit{Writing - Original Draft}: AP; \textit{Writing - Review \& Editing}: AP, SB, MJ, JB, RM, HL; \textit{Conceptualization}: AP, SB, MJ, JB, RM, HL; \textit{Investigation}: AP, SB; \textit{Methodology}: AP, SB; \textit{Software}: AP, SB; \textit{Supervision}: MJ, JB, RM, HL; \textit{Project Administration}: MJ, HL; \textit{Funding Acquisition}: MJ, JB, RM, HL.
\section{}

\bibliography{Bibliography}
\bibliographystyle{IEEEtran}

\end{document}